\documentclass[letterpaper]{article} 
\usepackage{aaai23}  
\nocopyright
\usepackage{times}  
\usepackage{helvet}  
\usepackage{courier}  
\usepackage[hyphens]{url}  
\usepackage{graphicx} 
\usepackage{natbib}  
\usepackage{caption} 
\usepackage{algorithm}
\usepackage{algorithmic}

\usepackage{amsmath,amsfonts,amssymb}
\usepackage{spverbatim}
\usepackage{verbatimbox}
\usepackage{newfloat}
\usepackage{listings}
\DeclareCaptionStyle{ruled}{labelfont=normalfont,labelsep=colon,strut=off} 
\floatstyle{ruled}
\newfloat{listing}{tb}{lst}{}
\floatname{listing}{Listing}
\usepackage{color}

\newcommand\blfootnote[1]{%
  \begingroup
  \renewcommand\thefootnote{}\footnote{#1}%
  \addtocounter{footnote}{-1}%
  \endgroup
}

\title{Towards Long-term Autonomy: A Perspective from Robot Learning}
\author{
    Zhi Yan\textsuperscript{\rm 1}\blfootnote{This work was supported by the Bourgogne-Franche-Comt\'e regional research project LOST-CoRoNa and the Czech Science Foundation research project number 20-27034J 'ToltaTempo'. Nicola Bellotto has received funding from the EU Horizon 2020 research
and innovation programme under grant agreement No 101017274 (DARKO).},
    Li Sun\textsuperscript{\rm 2},
    Tomas Krajnik\textsuperscript{\rm 3},
    Tom Duckett\textsuperscript{\rm 4},
    Nicola Bellotto\textsuperscript{\rm 4,5}
}
\affiliations{
    \textsuperscript{\rm 1}CIAD UMR 7533, Univ. Bourgogne Franche-Comt\'e, UTBM, F-90010 Belfort, France\\
    \textsuperscript{\rm 2}Department of Computer Science, University of Sheffield, UK\\
    \textsuperscript{\rm 3}Faculty of Electrical Engineering, Czech Technical University, Czechia\\
    \textsuperscript{\rm 4}Lincoln Centre for Autonomous Systems, University of Lincoln, UK\\
    \textsuperscript{\rm 5}Department of Information Engineering, University of Padua, Italy
}

\begin{document}

\maketitle

\begin{abstract}
  In the future, service robots are expected to be able to operate autonomously for long periods of time without human intervention.
  Many work striving for this goal have been emerging with the development of robotics, both hardware and software.
  Today we believe that an important underpinning of long-term robot autonomy is the ability of robots to learn on site and on-the-fly, especially when they are deployed in changing environments or need to traverse different environments.
  In this paper, we examine the problem of long-term autonomy from the perspective of robot learning, especially in an online way, and discuss in tandem its premise ``data'' and the subsequent ``deployment''.
\end{abstract}

\section{Introduction}

Mobile robotics has come a long way in the past three decades, including the introduction of probabilistic methods with great success in dealing with the uncertainty during robot operation~\cite{pr}.
As the fundamental problems of robot autonomy are being solved one by one, we are looking forward to the \emph{long-term autonomy} of robots, while one of the important enabling technologies for the latter is machine learning~\cite{Kunze2018}.
Today, an active community bridging machine learning and robotics research and technology 
is growing, also known as \emph{robot learning}, to address the unique challenges posed by the intersection of these two disciplines, such as:
\begin{enumerate}
\item How to obtain useful (high-quality) and sufficient (exhaustive) sensory data through robot perception?
\item How to deal with the changes in the environment as well as different environments?
\item How to ensure the real-time performance of the integrated system under the limited on-board computing resources of the robot?
\end{enumerate}
In this paper, we conduct some discussions on these three aspects, based on our research experience in this domain.

\section{About Data}

The acquisition of perceptual data mainly relies on various sensors, both for proprioception and exteroception.
The former includes wheel encoders, IMU, thermometer, etc., while the latter includes lidar, camera, sonar, etc.
From a long-term autonomy perspective, proprioceptive data are often easy to interpret and model, since the objects they measure are inherently invariant. 
An intuitive example is battery health monitoring.
The exteroceptive data are more complex, mixing order and disorder,  because environmental changes may feature both regularity~\cite{fremen} and randomness~\cite{ls18icra}.
For example, you may know that a particular corridor will be crowded at a certain time, but it is hard to predict the exact trajectories of people going through it~\cite{tomek,vintr2020natural}.

Exteroceptive data have more impact on long-term robot autonomy, as they close the robot-world feedback loop, and different sensor modalities have their own advantages and disadvantages.
For example, cameras can provide color and texture information to help capture rich environmental semantics, but are affected by lighting conditions.
Lidars can usually cope with the latter, but the data generated is a set of coordinate points in 3D space, and its sparseness increases with the measurement distance, making it difficult to interpret.
Furthermore, in outdoor scenarios, long-term robot autonomy will inevitably have to deal with the challenges posed by various weather conditions. 
These are under-represented in many datasets used in machine learning, e.g., out of the 328 000 pictures in the COCO dataset~\cite{10.1007/978-3-319-10602-1_48}, 
only one shows foggy conditions~\cite{broughton2022}.
However, in foggy conditions, cameras and lidars are significantly affected. 
A popular solution is to substitute cameras and lidars with millimeter-wave radars, but interpreting radar data can be difficult~\cite{3l4av}.

Therefore, to answer our Question 1, multimodal sensor schemes are a direction worthy of vigorous research.
However, multimodal information fusion may be more complicated than imagined.
A realistic industrial case is Tesla's announcement earlier this year that it will completely switch to a purely visual vehicle perception solution~\cite{tesla}.
In addition to a small part of the hardware cost considerations, the main reason is that the technical cost of multimodal information fusion is really incalculable.
Certainly, one needs to note that the reason why Tesla has the confidence to stick to the pure vision solution is related to the number of its car users, which is the source of the large amount of learning data that the former relies on.

Existing fusion schemes can be roughly divided into low-level data fusion and high-level information fusion.
An example of the former is pixel-to-point matching of images and point clouds, while an example of the latter is fusion of detection results of different object detectors.
Obviously, high-level fusion is accompanied by a higher demand for AI.
For example, how should we choose when different detectors give different results for the same object?
As an example, \cite{yz18iros} applied the odds, given that each detector can provide its detection confidence (expressed as a probability):
\begin{equation}
P(Y | \{ d^j_i \}) = \dfrac{odds(Y | \{ d^j_i \})}{1 + odds(Y | \{ d^j_i \})}
\end{equation}
where
\begin{equation}
odds(Y | \{ d^j_i \}) = \prod^{o}_{j=1} \prod^{k}_{i=1} odds(Y | d^j_i)
\end{equation}
and
\begin{equation}
odds(Y | d^j_i) = \dfrac{P(Y | d^j_i)}{P(\neg Y | d^j_i)} = \dfrac{P(Y | d^j_i)}{1 - P(Y | d^j_i)}.
\end{equation}
where $P(Y)$ represents the probability of belonging to object $Y$, and $d^j_i$ represents the $i$-th detection result provided by the $j$-th detector.
This fusion approach proves to be a good compromise when dealing with inconsistent information as it takes into account the potential interactions among multiple sensors~\cite{moravec88sensor,stepan} and implicitly considers time~\cite{yz18iros}.

On the other hand, passively acquiring sufficient amount of useful data usually comes with a time cost.
However, mobile robots can actively explore the environment for useful observations, thus providing strong technical support for effectively acquiring high-quality and sufficient data~\cite{Santos2016,Santos2017,ferdi2022,sergi2020}.
For example, an end-to-end method for active object classification with RGB-D data is proposed in~\cite{patten16ral}, which plans a robot's future observations to identify uncertain objects in clutter.
Their results show that the proposed active method can recognise objects faster and effectively reduce the number of views required to accurately classify cluttered objects compared to traditional passive perception.

\section{About Learning}

Some changes in the environment are fast (e.g. pedestrians), some are slow (e.g. layouts), some are regular (e.g. day and night) while some are random (e.g. human clothing)~\cite{Biber2009,Krajnik2016}.
In addition, different environments may not only have different appearances and layouts, but the behavior of the humans in them may also differ.
For example, people's pace in a supermarket and a hospital can be very different due to their different purposes and moods.
In theory, no static model can fully adapt to changes in the environment and different environments.
Therefore, models of dynamics than represent the observed changes in their temporal context are envisaged.
Like humans, this can be done in a learn-by-use fashion.

\begin{figure}[t]
  \centering
  \includegraphics[width=\columnwidth]{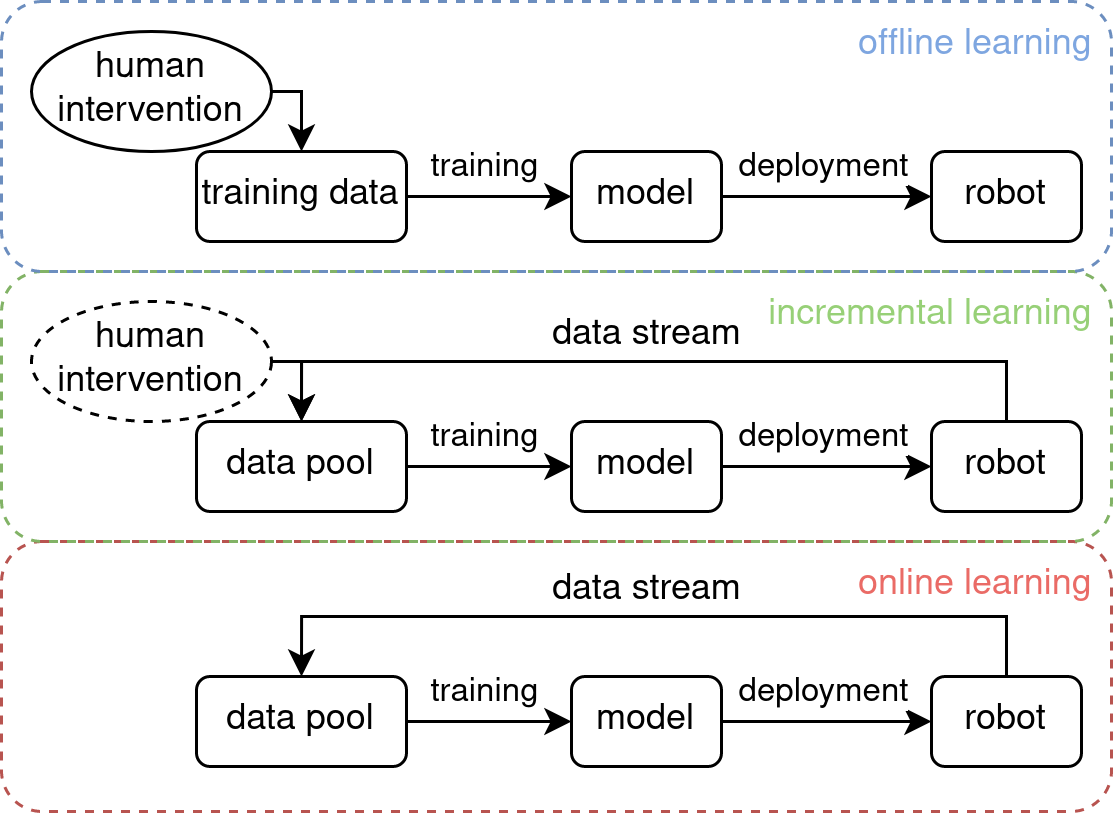}
  \caption{The use of three different learning paradigms in mobile robotics.}
  \label{fig:learnings}
\end{figure}

In our research, we emphasize the \emph{in-situ and on-the-fly} learning ability of robots, methodologically calling it online (machine) learning.
Generally speaking, online learning is a method of machine learning in which data becomes available in a sequential order and is used to update the best predictor for future data at each step, as opposed to batch learning techniques that generate the best predictor by learning on the entire training data set at once.
A typical (industrial) application of online machine learning is the prediction of user preferences by Internet companies, such as accurate advertisement placement, product recommendation, and so on.
In the field of mobile robotics, our research put more emphasis on ``autonomy'' for the ``online'' aspect, that is, robots should learn spontaneously and automatically without human intervention.
In order to clarify this, we take offline learning and incremental learning as references, and further interpret online learning through a comparative illustration.
The intuitive difference between the three can be seen in Figure~\ref{fig:learnings}.

Offline learning, similar to offline programming commonly found in industrial robotic arms, refers to the fact that a model is fully trained before deployment to the robot and remains unchanged for the duration of the robot operation.
To this end, the data is collected in advance and usually annotated in order to guarantee the final obtained model performance.
Incremental learning can be online or offline.
It can handle continuous data without imposing timeliness.
It also allows human intervention when necessary to ensure the iterative learning performance of the model.
This way of learning places more emphasis on the preservation of knowledge and the avoidance of catastrophic forgetting.
Instead, online learning means that the robot learns autonomously during its operation without human intervention at all.
Timeliness is another important feature of this learning method, that is, learning quickly and applying the learned model immediately.

\subsection{Formulation of online learning}

Suppose that the total number of data we can finally obtain is $D$, but at a certain time $t$ we can only obtain data from $1$ to $t-1$, denoted as:
\begin{equation}
  D_{t-1} = (x_1, y_1), (x_1, y_1), \dots, (x_{t-1}, y_{t-1})
\end{equation}
Online learning is designed to predict the corresponding output $\hat{y}=h(x_t,w_t)$
for the input $x_t$ at the current moment based on the data that has been observed $D_{t-1}$, where $w_t$ is the set of model parameters.
Since we cannot obtain all the data at time $t$, we hope that the decisions made at each time are close to the decisions made when all the data are finally obtained.
Therefore, the optimisation objective of online learning can be defined as:
\begin{equation}
  Regret(T) = \sum\limits_{t=1}^T l(x_t,y_t,w_t) - \sum\limits_{t=1}^T l(x_t,y_t,w_*)
\end{equation}
where $w_* = argmin_w\sum_{t=1}^Tl(x_t,y_t,w_t)$ refers to the model parameters obtained after getting all the data, and $l$ is a (positive) loss function.
Obviously, this is unlikely to happen for long-term robot autonomy.
In addition, a unique challenge that needs to be faced in the field of robotics is that incoming data is often unannotated.
We look at the former problem from the perspective of online stability in the following section, considering the situation of incomplete data. Unsupervised~\cite{Ambrus2015}, self-supervised~\cite{yz19auro} and reinforcement~\cite{DelDuchetto2022} learning schemes help to solve the latter problem. We further discuss our solutions using self-supervised learning in some actual case scenarios.

\subsection{Convergence and stability}

One of the challenges of online learning is how to converge quickly without human intervention and maintain stable system performance over time.
We formalize the learning stability (equivalent to dynamic regret~\cite{zinkevich03icml}) as:
\begin{equation}
  Stability(T) = \sum\limits_{t=1}^T \| u_t - u_{t+1} \|
\end{equation}
where $u_t$ is the number of correct predictions by the model at time $t$.
While according to Lyapunov stability, the iterative system will stabilize if:
\begin{equation}
  \lim_{T \to \infty}\frac{Stability(T)}{T} = 0
\end{equation}

\begin{figure}[t]
  \centering
  \includegraphics[width=\columnwidth]{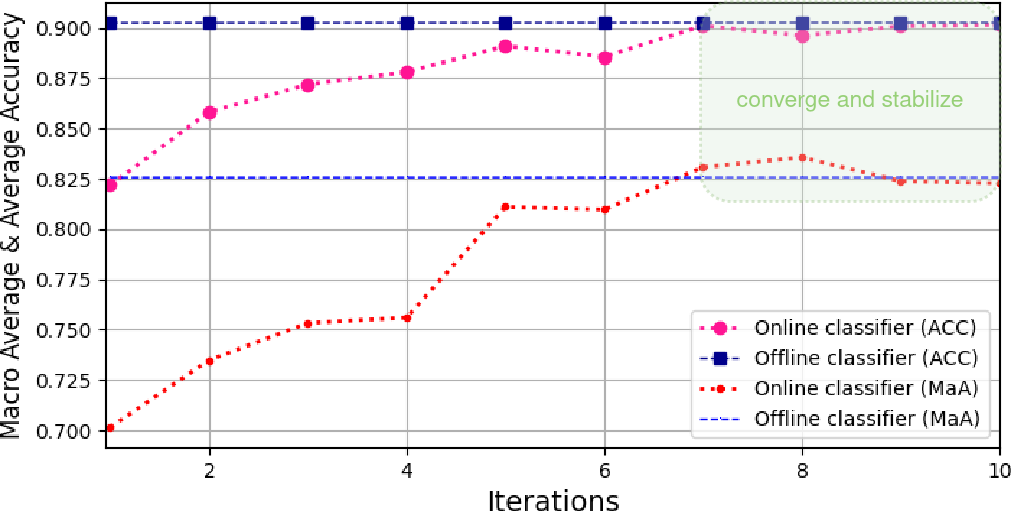}
  \caption{Convergence and stability analysis of online learning.}
  \label{fig:convergence}
\end{figure}

A simple practice is to set corresponding performance expectations based on some established metrics (such as determined by offline trained baseline methods~\cite{yang21itsc}, see Figure~\ref{fig:convergence}), and then perform online learning iterations based on these.
Once the performance of the model learned online falls within the tolerance range of the expected value over a period of time, we consider the model performance to have stabilized.
In addition, how to construct the test set, how to avoid catastrophic forgetting, etc. to ensure long-term stability also need to be considered~\cite{lesort20if}.

\subsection{Some actual case scenarios}

\begin{figure}[t]
  \centering
  \includegraphics[width=\columnwidth]{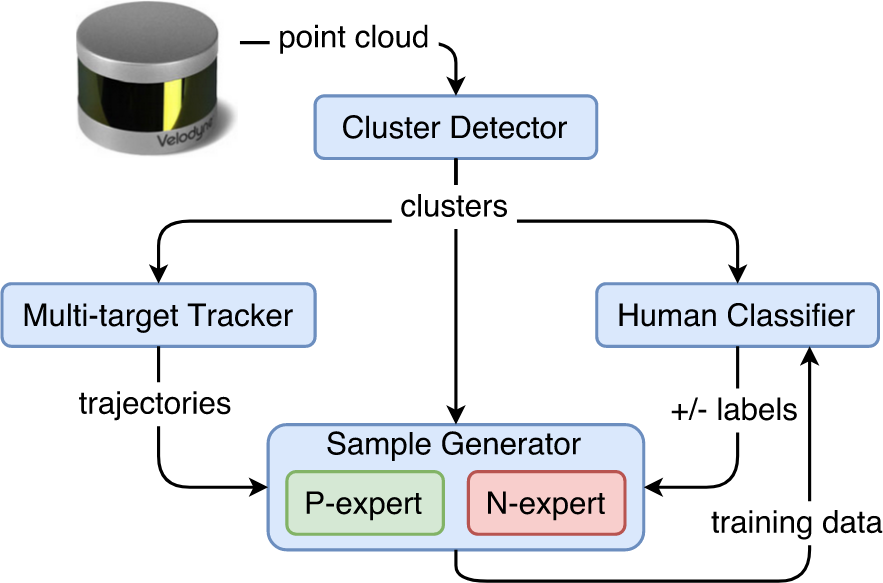}
  \caption{Online learning for human classification in 3D lidar-based tracking.}
  \label{fig:yz17iros}
\end{figure}

In~\cite{yz17iros}, an online learning framework is proposed for robots to rapidly learn a human detector for lidar point clouds in their deployed environment.
As shown in Figure~\ref{fig:yz17iros}, the point cloud of lidar is first input to a clustering module, and the output of the segmented point clusters is respectively input to a multi-target tracker and a human classifier that can be dynamically updated online.
The multi-target tracker is responsible for correlating multiple observations belonging to the same target to generate its trajectory.
Through a pair of rule-based heuristic experts -- a P-expert responsible for avoiding false negatives and an N-expert responsible for eliminating false positives -- high-quality learning samples are continuously fed to a human classifier for learning.
These ``experts'' can be based on object movement distance, average speed, tracking covariance, and more.
It is worth noting that since online learning is sensitive to the correctness of early learning samples, the proposed method has to resort to human-supervised samples during initial learning.
However, it aims to reduce this dependence to a minimum, i.e. a single sample is also feasible.

\begin{figure}[t]
  \centering
  \includegraphics[width=\columnwidth]{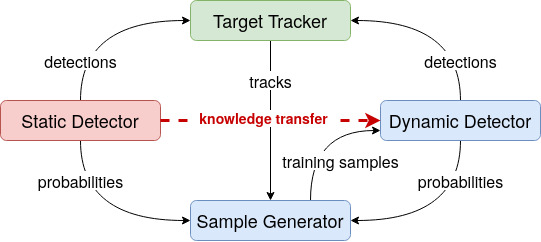}
  \caption{Multisensor online transfer learning for 3D lidar-based human detection.}
  \label{fig:yz18iros}
\end{figure}

It can be seen that the above method still requires human intervention in the initial stage of learning, and the discriminative power of rule-based heuristic experts is limited.
Therefore, in~\cite{yz18iros}, a further approach is proposed: using one sensor to train a different one, or more broadly, one model continuously learning from another one, as shown in Figure~\ref{fig:yz18iros}.
This improved framework consists of two detectors: the static one, which was already trained, and the dynamic one, which the robot needs to learn continuously during its operation.
The two are still linked by a multi-target tracker.
Specifically, they send the detected human locations to the tracker to generate corresponding trajectories, while also feeding the corresponding detection probabilities to a label generator.
The latter uses the odds-based probabilistic fusion method, mentioned earlier, to annotate a set of samples associated with human trajectories and feed them to the dynamic detector for learning.
It can be seen that the improved online learning framework eliminates the need of human intervention during the robot's operation, and also it enables the ability of the experts (static detectors) to be maintained.

\section{About Deployment}

A strong feature in the field of robotics is the real-time requirement.
This involves many aspects such as the integration of software and hardware, the development of algorithms, the implementation of codes, and so forth~\cite{Hawes2017,Biswas2016,Duchetto2019}.
Real-time systems can be divided into soft real-time (e.g. ROS) and hard real-time (e.g. ROS2).
The former gives a certain tolerance to the response of the system, while the latter forces the process to complete in a very short time, otherwise the system is considered to have failed.

Standing at the intersection of robotics and machine learning, combined with our experience in software engineering, we argue that ensuring the real-time nature of robotic learning systems can be approached from the following aspects:
\begin{itemize}
\item Distributed deployment. An inherent property of ROS is native support for distributed computing. Judging from the current state of hardware development, multiple computing units are still the mainstream solution for robotic systems, especially when it comes to requiring GPU computing power. Accordingly, it is a good practice to adapt the software to its suitable hardware, such as deploying system security and control-related components to the CPU side, while deploying heavy visual computing or deep neural networks to the GPU side.
\item Algorithm optimization. In a world where data-driven solutions are popular, we are still willing to focus on some basic properties of algorithms, such as complexity. The complexity is divided into time complexity and space complexity. For robot learning, one may be concerned about the time complexity of model training and prediction, and the space complexity of the runtime after the model is deployed. Taking Support Vector Machines (SVM) as an example, the time complexity of model training varies with different kernels, ranging from $O(n^2)$ to $O(n^3)$, where $n$ represents the number of training samples. Its prediction time complexity ranges from $O(f)$ to $O(sf)$, where $f$ represents the number of features and $s$ represents the number of support vectors. In general, linear kernels have $O(f)$, while RBFs and polynomials have $O(sf)$. The runtime space complexity of SVM is $O(s)$.
\item Implementation of the code.
As shown in, e.g.,~\cite{fivser2013growing}, proper programming practices can speed up machine learning methods by several orders of magnitude.
The basic principles are to avoid useless loops, try not to create variables within them, try to pass references instead of values, and so on.
The last point concerns the problem of avoiding data replication.
ROS-related practices include using nodelets instead of nodes.
A concrete example is that one of the reasons for the improvement in data clustering performance reported in~\cite{yz19auro} relative to that reported in~\cite{yz17iros} is to replace

\begin{verbnobox}[\fontsize{8pt}{8pt}\selectfont]
pcl::PointCloud<pcl::PointXYZI>::Ptr
clusters(new pcl::PointCloud <pcl::PointXYZI>);
\end{verbnobox}
with
\begin{verbnobox}[\fontsize{8pt}{8pt}\selectfont]
std::vector<pcl::PointCloud <pcl::PointXYZI>::
Ptr, Eigen::aligned_allocator <pcl::PointCloud
<pcl::PointXYZI>::Ptr > > clusters;
\end{verbnobox}
thus avoiding data duplication.

\end{itemize}

\section{Conclusions}
In this paper, we look at the problem of long-term robot autonomy from the perspective of robot learning, an emerging field that straddles mobile robotics and machine learning.
Our elaboration mainly comes from three aspects including data acquisition, online learning, and engineering deployment.
Our insights are neither comprehensive nor exclusive, but only a rough summary based on our hands-on experience, hoping to provide some reference for the community and play a role in attracting more ideas.
\pagebreak{}

\bibliography{ref}

\end{document}